# LLMs for Legal Subsumption in German Employment Contracts

Oliver Wardas
oliver.wardas@tum.de
Technical University of Munich
Munich, Bavaria, Germany

Florian Matthes
matthes@tum.de
Technical University of Munich
Munich, Bavaria, Germany

## Abstract

Legal work, characterized by its text-heavy and resource-intensive nature, presents unique challenges and opportunities for NLP research. While data-driven approaches have advanced the field, their lack of interpretability and trustworthiness limits their applicability in dynamic legal environments. To address these issues, we collaborated with legal experts to extend an existing dataset and explored the use of Large Language Models (LLMs) and in-context learning to evaluate the legality of clauses in German employment contracts. Our work evaluates the ability of different LLMs to classify clauses as "valid," "unfair," or "void" under three legal context variants: no legal context, full-text sources of laws and court rulings, and distilled versions of these (referred to as examination guidelines). Results show that full-text sources moderately improve performance, while examination guidelines significantly enhance recall for void clauses and weighted F1-Score, reaching 80%. Despite these advancements, LLMs' performance when using full-text sources remains substantially below that of human lawyers. We contribute an extended dataset, including examination guidelines, referenced legal sources, and corresponding annotations, alongside our code and all log files. Our findings highlight the potential of LLMs to assist lawyers in contract legality review while also underscoring the limitations of the methods presented.

## 1 Introduction

The text-based and resource-intensive nature [7, 14] of legal work has made it a compelling area for NLP research over the past five decades, with interest continuing to grow [15]. Initiated by Buchanan and Headrick in 1970 [4], early legal NLP efforts were dominated by symbolic approaches aimed at formalizing legal knowledge to enable automated legal reasoning [27]. However, the inherently ambiguous nature of law presents significant challenges to reasoning in a mathematical manner [25, 33]. This, together with the difficulty of acquiring explicit legal knowledge, have likely contributed to a recent shift in legal NLP research towards data-driven approaches that do not rely on explicit legal knowledge representation or reasoning [27]. While these purely statistical methods are less affected by the challenges of legal ambiguity, they come at the cost of explicit legal reasoning and may base their predictions on legally irrelevant (but predictive) or even illegal (e.g. ethnicity) factors [29]. Consequently, these approaches lack interpretability, transparency, and, by extension, trustworthiness. Additionally, legal frameworks evolve continuously with changing legislation and new court rulings. This continuous evolution of legal frameworks poses a significant challenge for purely data-driven methods. Models which get their predictive ability encoded into their weights quickly become outdated. The need for constant retraining to incorporate these changes renders such approaches impractical for real-world applications. Furthermore, existing training data itself is compromised over time, as it may contain precedents and reasoning that have been overruled or superseded. These fundamental issues undermine the reliability and utility of data-driven approaches in dynamic legal environments. As argued by Santosh et al. (2024) [27], combining data-driven approaches with explicit legal expertise holds promise to combine the strengths of both methods, enabling effective, yet interpretable, and trustworthy legal NLP systems. Modern Large Language Models have proven their ability to achieve state of the art results on academic benchmarks leveraging in-context learning [1]. It is to be expected that these capabilities will further increase with growing model sizes [3]. We believe that this development poses major opportunities to advance the field of legal NLP. In collaboration with legal experts, we extended an existing dataset and explored knowledge-based methods, leveraging Large Language Models (LLMs) and in-context learning, to evaluate the legality of clauses in German employment contracts. We therefore conducted a series of experiments, testing the ability of different LLMs to subsume clauses as void under different variants of legal knowledge, comprised of laws and court rulings. A clause had to be classified as "valid", "unfair" or "void" given either no legal context, full-text sources of relevant laws and court rulings, or distilled versions of these sources (examination guidelines), created by human lawyers. Findings show that while full-text sources moderately improve prediction performance, examination guidelines significantly enhance performance, achieving high recall for void clauses and over 80% weighted F1-Score. This demonstrates the potential of LLMs to assist lawyers in the given task. However, the results also show that the ability of LLMs to subsume under legal sources in the form in which they are released and available to the public lags behind human lawyers significantly. This finding is contrary to the results observed in studies of GPT-4 on the bar exam, which demonstrated strong performance compared to human lawyers [13]. Alongside classification metrics for GPT-4o, DeepSeek-V3, Grok-2, Gemini-1.5, and DeepSeek-R1, we release a new dataset extending the existing clause dataset used in this research. For each void clause, we include the corresponding examination guideline and references to the relevant laws and court rulings as well as full-text versions of these sources. To ensure full reproducibility, the code to run the experiments is also published as well as all generated log files[1].

## 2 Background

### 2.1 Employment Contracts

Employment contracts are legally binding agreements that delineate the mutual rights and obligations between employers and employees, establishing the foundational framework of the employment relationship. Such contracts encompass clauses addressing key elements such as remuneration, working hours, job duties, termination conditions, and other essential aspects of employment. A clause, is a semantically self contained component of the contract that governs a specific facet of the relationship, such as the work schedule. These clauses are usually grouped into broader sections, such as those covering execution, compensation, or termination. In this work, we predict the legality of individual clauses, as they typically represent the smallest adjudicable unit in legal disputes. Employment contracts ensure transparency and fairness between the two parties, thus safeguarding employees' rights and providing reliability for employers. Void clauses tend to pose unfair disadvantages for the employee and put the employer at risk of expensive lawsuits. Legal reviews of employment contracts therefore benefit both parties. Their central role in labor relations and the high cost of manual legal reviews make them a viable application for legal NLP to decrease costs and improve access to justice [5].

### 2.2 Legal Subsumption

Legal subsumption entails the systematic evaluation of whether a specific case or contractual provision aligns with the broader criteria established by laws and court rulings. This process requires meticulous interpretation of these legal documents, coupled with the analysis of relevant factual circumstances to determine whether they fulfill the conditions prescribed by the law. Legal practitioners rely on their education and expertise to navigate the complexities and ambiguities of laws, drawing upon precedents and contextual knowledge to render informed judgments. In this work, we seek to emulate this subsumption process by leveraging LLMs and in-context learning.

## 3 Related Work

### 3.1 Contract Legality Analysis

Although academic interest in Legal NLP research has grown significantly in recent years, Contract Legality Analysis, particularly in non-English languages, remains underexplored [15], possibly due to data scarcity. The majority of contracts are created between private entities, such as individuals or companies, and contain sensitive information, making them rarely accessible for research. In contrast, the task of judgment prediction, which utilizes data published by courts, has gained much more popularity. Prior research in technology-assisted contract legality analysis has explored rule-based approaches and machine learning. Lee et al. (2019) [16] developed a rule-based system to identify hazardous clauses in FIDIC-standard construction contracts. Hendrycks et al. (2021) [10] introduced a comprehensive benchmark dataset containing approximately 13,000 annotated clauses across 25 contract types, focusing on operational and financial risks. Similarly, Leivaditi et al. (2020) [17] compiled a dataset of 179 leasing contracts, identifying 17 risky clause types and training a machine learning model for clause classification. Research efforts within the CLAUDETTE project (Lippi et al., 2019) [18] and by Braun et al. (2021) [2] employed machine learning techniques to evaluate general terms and conditions in English and German online shop contracts. Wardas et al. (2025) [35] published the dataset we are using in this work alongside with baseline model evaluations, making it the only research effort on German employment contracts.

### 3.2 Legal Reasoning using NLP

Recent work in legal NLP included studies on various aspects of legal reasoning, particularly in the application of Legal Judgment Prediction. Yu et al. (2023) [37] and Parizi et al. (2023) [24] studied prompt engineering for legal tasks. Shu et al. (2024) [31] proposed *LawLLM*, a model for U.S. legal applications, outperforming baselines in Similar Case Retrieval, Precedent Case Recommendation, and Legal Judgment Prediction through in-context learning. Santosh et al. (2024) [28] and Nigam et al. (2023, 2024) [21, 22] integrated precedents and statutes to predict outcomes in cases from the European Court of Human Rights and the Supreme Court of India. Valvoda et al. (2024) [34] addressed reasoning in legal prediction models. Goodson et al. (2023) [9] applied language models for intention and context elicitation in legal aid. Schumacher et al. (2024) [30] analyzed argument reasoning in civil procedures using GPT-4 [1]. Huber-Fliflet et al. (2024) [12] explored in-context learning and Retrieval Augmented Generation for legal document review in construction delay disputes.

## 4 Data

Our main source of data is the dataset[2] published by Wardas et al. (2025) [35]. This dataset comprises 1094 samples, each containing a clause from a German employment contract, its section title, and three annotations. These annotations include: a legality class ("valid", "unfair", or "void"); one of 14 categories; and, if the clause is deemed unfair or void, a brief explanation outlining the reason. The definitions of the classes are as follows:

- **Valid (0):** A clause is valid if it is neither unfair nor void.
- **Unfair (0.5):** A clause is unfair if it creates significant disadvantages for one party without being definitely void.
- **Void (1):** A clause is void if this can be subsumed under existing law and court rulings.

---
[1]Employment Contract Dataset V2

[2]Employment Contract Dataset V1

In this work, we extend the given dataset (a filtered version where we removed duplicates, keeping only the most critical annotation) by a set of examination guidelines - distilled legal rules under which certain groups of clauses can be subsumed as void. Each examination guideline is associated with one of the 14 clause categories [35] and contains references to legal sources from which it was derived. Those full-text legal sources are also included. Every void clause is annotated with the examination guideline under which it can be subsumed as void. Table 1 shows an example of such an examination guideline and a clause which would be subsumed void under it.

Table 1: Example Examination Guideline

| | |
|---|---|
| Content | Eine Klausel, die in den AGB ein generelles Pfändungs- oder Abtretungsverbot für die Forderungen des Arbeitnehmers festlegt, ist unwirksam. |
| Content (en) | A clause that stipulates a general prohibition of attachment or assignment of the employee's claims in the GTC is void. |
| Topic | Pfändung/Abtretung |
| Topic (en) | Assignment/Garnishment |
| References | § 308 Nr. 9 lit. a BGB; BGH, Urteil vom 1. Februar 1978 (VIII ZR 232/75) |
| Void Clause | Die teilweise oder vollständige Abtretung und/oder Verpfändung der Vergütung an Dritte ist ausgeschlossen. |
| Void Clause (en) | The partial or complete assignment and/or pledging of the remuneration to third parties is excluded. |

### 4.1 Annotation Process

The examination guidelines were created by two lawyers from a German law firm, specialized in economic law. First, they grouped clauses together which were void for the same reason. Then they formulated one examination guideline for each group, describing the legal argument under which these clauses would be ruled void in court. In addition, they annotated each examination guideline with references to all the laws and court rulings required to derive the guideline.

### 4.2 Insights

The filtered dataset contains 891 samples from which 697 are classified as "valid", 82 as "unfair", and 112 as "void". In total, 24 examination guidelines were derived from a total of 41 laws and court rulings. Table 2 gives insights into the length of these documents. With a total of 112 void clauses in the dataset, every examination guideline covers a group of ~4.7 void clauses on average.

## 5 Experiments

The primary aim of this work is to assess the legal subsumption ability of Large Language Models (LLMs) under varying legal contexts. This evaluation focuses exclusively on measuring classification metrics. While legal reasoning, in the form of generated explanatory text supporting the classifications, remains outside the scope of this paper, our methodology assumes that the subsumption ability can be meaningfully quantified without it. However, all explanations were collected and are published for further research. The complexity of retrieval mechanisms (e.g. to find fitting legal full-text sources) are intentionally excluded. By assuming a perfect retrieval system, we isolate the models' subsumption ability as the sole variable of interest. Positive results from this research emphasize the need for further investigations into effective retrieval mechanisms in future work.

Table 2: Document Statistics

| | | Word Counts | | |
|---|---|---|---|---|
| Document Type | Count | Min. | Max. | Avg. |
| Examination Guidelines | 24 | 10 | 83 | 27 |
| Laws | 12 | 17 | 563 | 160.25 |
| Court Rulings | 29 | 126 | 8660 | 3722.66 |

### 5.1 Model Selection

To perform the challenging task of legality analysis, we selected some of the highest-ranking models in the German category of the Chatbot Arena LLM Leaderboard [6] (accessed: 21.01.2025). We excluded all models that did not meet the following criteria:

- **Context Size:** Models must provide a minimum context window of 64k tokens to accommodate multiple full-text legal sources, each potentially containing up to 8k characters, within a single prompt.
- **API Rate Limits:** Models must be available without restrictive rate limits to allow experiments to run in a reasonable time. This excluded *gemini-exp-1206* and *gemini-2.0-flash-exp*, which restricted access after 10 consecutive requests.
- **Structured Output:** Models must reliably return valid JSON-formatted output as described in Section 5.2, ideally enforceable through additional request parameters. This excluded *claude-3-5-sonnet-20241022*, which consistently failed to follow formatting instructions.

Additionally, to ensure diversity in our model selection, we only included the highest-ranking model from each provider that met all the above criteria. The following models were selected (ordered by their ranking):

**GPT-4o**, developed by OpenAI [1], is the top-ranking model in the German category of the leaderboard. GPT-4, first released in 2023, gained significant attention for its outstanding performance in legal contexts, particularly its near 90th percentile score on the United States bar exam [13]. For our experiments, we utilize the latest version, *gpt-4o-2024-11-20*, released on 2024-11-20.

**DeepSeek-V3**, created by DeepSeek AI [19], is one of the two open-source models included. It represents the latest iteration in the DeepSeek series and was released on 2024-12-27. This model is notable for its cost-efficient training, reportedly requiring only 2.788 million H800 GPU hours across all training stages.

**Grok** is a Large Language Model family developed by xAI [36]. For our experiments, we use *grok-2-1212*, released on 2024-12-14.

**Gemini**, a family of Large Language Models by Google [32], is also part of our selection. Although more recent versions exist, we chose *gemini-1.5-pro-002*, released on 2024-09-24. Newer models, such as *gemini-exp-1206* and *gemini-2.0-flash-exp*, were excluded due to restrictive rate limits.

**DeepSeek-R1**, the inaugural reasoning model from DeepSeek [8], was released on 2025-01-20. It is reportedly on par with OpenAI's *o1-2024-12-17* model [23] in terms of performance while being both open-source and significantly less expensive. Despite not yet being ranked on the leaderboard due to its recent release, we included it as the representative reasoning LLM in our experiments.

## 5.2 Setup

All models were used over their APIs, provided by OpenAI, DeepSeek AI, xAI, and Google respectively. However, since DeepSeek-V3 and DeepSeek-R1 are open source, they can also be run locally. All APIs share a similar request body structure, consisting of a chain of existing messages, a system instruction to define the LLMs task (for the OpenAI, DeepSeek AI, and xAI API this is the first message) and additional parameters. These include options to specify and enforce the desired output format and a seed parameter (not supported for DeepSeek-V3 and DeepSeek-R1) to ensure reproducibility. The system instruction was used to describe the task, explain the three possible classes, provide additional context like examination guidelines or full-text legal sources (laws and court rulings), and the desired output structure (full prompts in the appendix A.1). The only user message was the clause text itself. For all LLM responses, the following output structure was requested and enforced:

```
{
  "explanation": <explanation for classification>,
  "result": <classification>,
  "hurt_rules": [<examination guideline id>]
}
```

The "hurt_rules" field is only used when providing examination guidelines to the model. For the two variants in which legal context is given to the models, they are instructed to only assess clauses as void if this can be subsumed under the given context. This aims to override any preexisting internal legal knowledge of the LLMs which may be wrong, outdated or incomplete.

## 5.3 Legal Context Variants

For each of the models, we compared three different variants of prompts and legal context, leading to a total of 15 (-1) executions over the entire dataset. The only exception is DeepSeek-R1 which could not be used with full-text sources due to long inference times and unreliably output formatting.

**No Legal Context:** The simplest variant is to provide no legal context at all. Since the models do not perform internet searches when used over the API, they have to solely rely on their internal knowledge. This variant serves as the baseline performance for each model. It is expected that the internal legal knowledge (especially for German law) is very limited and potentially outdated; providing any kind of legal context should yield better results.

**Examination Guidelines:** In this variant, all examination guidelines are provided for every clause evaluated, without any filtering. This approach is feasible because the guidelines are concise and short, eliminating the need for pre-selection. By avoiding filtering mechanisms, we directly test the models' ability to identify which guidelines are applicable to a given clause. While filtering based on clause topics or embeddings might be practical for larger knowledge bases, it introduces additional complexity and potential inaccuracies. Our approach ensures that results solely reflect the models' subsumption capabilities and are not distorted by potentially imperfect filtering.

**Full-Text Sources:** In this variant, the court rulings and laws provided to the models are filtered to simulate a perfect retrieval system. This ensures that if a clause is void, the sources include the exact references under which it can be subsumed as void. To achieve this, we utilize the annotations linking void clauses to examination guidelines, which in turn reference the relevant sources. These specific sources are supplied to the models, ensuring that all necessary legal context is present. For clauses that are not void, we still provide some legal sources to prevent the models from achieving perfect precision by defaulting to "valid" classifications in the absence of context. In such cases, two sources are randomly selected, prioritizing those matching the clause's topic, by identifying examination guidelines relevant to the clause's topic and randomly choosing two of their references.

## 5.4 Results and Discussion

For all constellations of models and context variants, we measured precision, recall, and F1-Score for each class and the averaged F1-Score (macro and weighted). The only exception is DeepSeek-R1, whose inference time and unreliably output formatting forced us to skip this constellation. Table 3 shows the results and highlights the best performing models for each measurement and context variant.

**General Observations:** The first general observation is that providing legal context in any form enhances the prediction performance in both macro and weighted F1-Score for every model evaluated. This underlines the importance of high quality legal knowledge bases and retrieval mechanisms for legal NLP research and applications. Secondly, examination guidelines proved to be more effective than full-text sources. This shows that while LLMs are capable of performing legal subsumption to a decent degree when provided with distilled legal knowledge, preprocessed by human lawyers, their capabilities in applying laws and court rulings directly, lack behind human lawyers significantly. The fact that these experiments were carried out undistorted under the assumption of a perfect retrieval system makes clear that current state of the art LLMs are still too limited in their legal subsumption ability to carry out unassisted legality reviews for German employment contracts based on original legal sources.

Table 3: Classification Metrics

|  | Valid | | | Unfair | | | Void | | | Averaged F1 | |
| --- | --- | --- | --- | --- | --- | --- | --- | --- | --- | --- | --- |
|  | Precision | Recall | F1 | Precision | Recall | F1 | Precision | Recall | F1 | Macro | Weighted |
| **No Legal Context** | | | | | | | | | | | |
| gpt-4o-2024-11-20 | 0.87 | **0.76** | **0.81** | **0.18** | 0.46 | **0.26** | **0.42** | 0.27 | **0.33** | **0.46** | **0.70** |
| DeepSeek-V3 | 0.88 | 0.61 | 0.72 | 0.15 | **0.54** | 0.24 | 0.30 | 0.32 | 0.31 | 0.42 | 0.62 |
| grok-2-1212 | 0.86 | 0.62 | 0.72 | 0.14 | 0.39 | 0.20 | 0.22 | 0.30 | 0.26 | 0.39 | 0.62 |
| gemini-1.5-pro-002 | 0.89 | 0.60 | 0.72 | 0.13 | 0.52 | 0.21 | 0.33 | 0.25 | 0.28 | 0.40 | 0.62 |
| DeepSeek-R1 | **0.94** | 0.27 | 0.41 | 0.05 | 0.09 | 0.06 | 0.19 | **0.91** | 0.31 | 0.26 | 0.37 |
| **Examination Guidelines** | | | | | | | | | | | |
| gpt-4o-2024-11-20 | 0.92 | **0.83** | **0.88** | **0.20** | 0.30 | 0.24 | **0.67** | 0.80 | **0.73** | **0.62** | **0.80** |
| DeepSeek-V3 | 0.89 | 0.81 | 0.85 | 0.19 | 0.29 | 0.23 | 0.56 | 0.65 | 0.60 | 0.56 | 0.76 |
| grok-2-1212 | 0.92 | 0.71 | 0.80 | 0.15 | 0.32 | 0.21 | 0.50 | 0.82 | 0.62 | 0.54 | 0.72 |
| gemini-1.5-pro-002 | 0.91 | 0.73 | 0.81 | 0.18 | 0.41 | **0.25** | 0.56 | 0.71 | 0.62 | 0.56 | 0.73 |
| DeepSeek-R1 | **0.96** | 0.58 | 0.72 | 0.15 | **0.48** | 0.23 | 0.51 | **0.98** | 0.67 | 0.54 | 0.67 |
| **Full-Text Sources** | | | | | | | | | | | |
| gpt-4o-2024-11-20 | 0.87 | **0.78** | 0.82 | 0.14 | 0.34 | 0.19 | 0.73 | 0.37 | 0.49 | 0.50 | 0.72 |
| DeepSeek-V3 | 0.89 | 0.77 | **0.83** | **0.19** | 0.49 | **0.27** | 0.73 | 0.49 | 0.59 | **0.56** | **0.75** |
| grok-2-1212 | 0.89 | 0.74 | 0.81 | 0.17 | 0.50 | 0.25 | **0.81** | 0.49 | **0.61** | 0.56 | 0.73 |
| gemini-1.5-pro-002 | **0.91** | 0.63 | 0.74 | 0.15 | **0.55** | 0.24 | 0.55 | **0.54** | 0.54 | 0.51 | 0.67 |

**Examination Guideline Performance:** Applications which involve human-in-the-loop approaches for maintaining a distilled legal knowledge base and proofread LLM-generated legality reviews of German employment contracts seem to be within reach and could improve efficiency by providing initial drafts. This claim is most supported by the high recall scores (>80%) for void clauses which were achieved by GPT-4o, Grok-2 and DeepSeek-R1. The latter one achieved a near perfect recall of 98% though at the expense of a lower precision score (51%) than other models. Considering the fact that false negatives for void clauses are the single biggest issue and risk (because of lawyer liability), it might be worth considering using DeepSeek-R1 over GPT-4o despite GPT-4o achieving higher F1-Scores on average.

**DeepSeek-R1:** When using examination guidelines, DeepSeek-R1's recall for void clauses is unmatched but the precision is on the lower end. Together with the comparably low recall for valid - and the high recall for unfair clauses, the suspicion rises that the model is not necessarily better at subsuming void clauses as void, but instead exhibits a stronger bias towards the employee side, viewing every clause more critically than other models. This assumption is strongly supported by DeepSeek-R1's performance under no legal context. The void class recall is 91%, while averaging around 30% for all other models. At the same time all other metrics (besides the valid class precision) are outstandingly low, especially notable with the recall for valid clauses being at a record low of 27%, leading to the low averaged F1-Scores. This shows the model's heavy bias towards void classifications, questioning its subsumption ability under legal context. Our measurements align with the findings of Parizi et al. (2023) [24], who showed that performance improvements achieved by applying Prompt Engineering approaches like Chain-of-Thought on generic tasks, do not easily translate to knowledge-intensive domains like law. Despite these issues, the observations make a good case for evaluating other reasoning LLMs on this task (e.g. OpenAI's o1 model [23]) and taking a deeper dive into their reasoning process and argumentation.

**The Unfair Class:** While the focus of this work lies on the void class, we have to acknowledge the poor prediction performance regarding the unfair class. Regardless of legal context variant, all models exhibit low recall and even lower precision scores. The results make it clear that none of the evaluated models comes close to reliably recognizing which clauses are unfair whilst not being void under the provided legal context. Neither examination guidelines nor full-text sources significantly increase the performance regarding this class; in fact they even lower it in some cases. The lines between unfair and void clauses are thin, especially because one can assume that void clauses will also treat some party unfairly. A lawyer's ability to identify unfair (but not void) clauses might stem from their practical experience rather than from scholarly knowledge. Future work could explore if few-shot learning can lead to improvements by showcasing a few similar clauses which have been classified as unfair in the past.

**Conclusion:** Concluding, one can say that GPT-4o is the best performing model across the board, achieving the highest averaged F1-Scores when using examination guidelines and no legal context. The concept of providing distilled legal knowledge instead of full-text sources has proven to be promising with current state of the art LLMs. This context variant yielded the highest averaged F1-Scores for all models except Grok-2, which performed slightly better using full-text sources but was still not the best in this category.

**GPT-4o Performance:** GPT-4o has the highest averaged F1-Scores when provided examination guidelines or no legal context. When working with full-text legal sources, it is outperformed by Grok-2 and DeepSeek-V3 - the hightest scoring model in that category. One differentiator when using full-text sources could be the LLMs ability to recognize and use important pieces of information in a large body of otherwise irrelevant text. This ability is measured by "Needle In A Haystack" (NIAH) [11] tests where the LLM is tested all across its context window. Both models have a context window of 128k tokens and both reportedly perform very well in that benchmark. The authors of DeepSeek-V3 [19] did not specify how exactly the test was executed, so we assume a single-needle test. GPT-4o reportedly achieves near perfect scores across different variants of the NIAH test [26]. The differentiating factor may therefore be somewhere else.

Figure 1 presents the confusion matrix for GPT-4o when using examination guidelines. Void clauses that are misclassified are more frequently placed in the unfair class rather than the valid class. Similarly, misclassified valid clauses tend to be assigned to the unfair class rather than the void class. This indicates that the model rarely makes entirely incorrect predictions (e.g. predicting valid as void or vice versa). Instead, its errors generally involve misjudging the fine boundary between valid/unfair and unfair/void. However, for clauses that are actually unfair, the model's performance falls below random guessing when accounting for class distribution. Given that reliably identifying void clauses is the most critical aspect of this task, the model's performance remains adequate for practical applications. Nonetheless, this limitation should be a focus of future research.

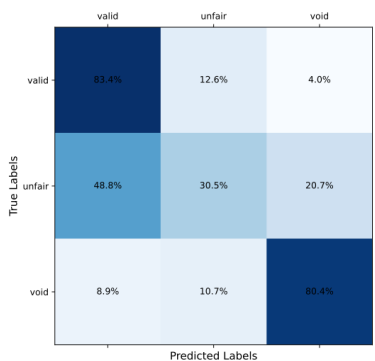

**Figure 1: Confusion Matrix GPT-4o**

## 6 Conclusion

### 6.1 Contribution

This work demonstrates that modern LLMs are capable of assisting with legality reviews of clauses in German employment contracts when provided with a knowledge base of distilled legal examination guidelines, derived from original sources such as laws and court rulings. However, when given the full-text of the original sources, LLMs struggle to subsume under them, indicating that they lag behind human lawyers in this ability. Despite this limitation, using full-text is still more effective than relying solely on the models' internal knowledge, highlighting the essential role of concepts like Retrieval Augmented Generation in applications using LLMs for legal tasks. These systems enable LLMs to leverage their text generation capabilities and general understanding to their full potential in solving legal tasks.

Beyond publishing the results of our experiments, we also provide an extended version of the clause dataset by Wardas et al. (2025) [35], which includes references to examination guidelines for void clauses, the guidelines themselves, references from these to legal documents such as laws and court rulings, and full-text versions of these documents (with 3 exceptions due to copyright restrictions from paid providers). The published data also contains all log files from our experiments, including the LLMs' explanations for their predictions. Furthermore, we release the complete code base used to conduct the experiments, ensuring full reproducibility.

### 6.2 Impact on Legal Practice and Processes

Reviewing contracts is one of the key tasks of lawyers and the technology used in this paper is readily accessible today. This work demonstrates that while LLMs have not yet achieved the full legal subsumption ability of human lawyers, they can provide meaningful assistance when supplied with the necessary legal knowledge in a distilled form (e.g. by creating initial drafts for lawyers to review). This opens opportunities for commercial applications to assist and accelerate lawyers in this task. Given the legal field chosen — employment contracts — the results of this work have wide-ranging relevance for companies, institutions, and individuals. This holds the promise of improving access to justice by reducing costs and lowering barriers for employees to use automatic or semi-automatic tools. Such advancements can significantly enhance fairness in the employer-employee relationship, promote transparency, and prevent legal disputes, thereby reducing the workload of courts.

### 6.3 Limitations

While this study reveals the potential and challenges of LLMs in assessing the legality of contractual clauses in German employment agreements by subsuming under a legal knowledge framework, it also faces several limitations. The full-text of the LAG Hessen decision from October 7, 2003 (13/12 Sa 1479/02) was unavailable, and a shorter summary was used instead. Legal sources were distributed across multiple portals, some of which were behind paywalls. They were available in different formats, such as HTML and PDF. One source was not even digitized and required OCR processing. Additionally, text formatting and annotations, such as headers containing metadata (e.g. referenced laws and court rulings), varied among different providers. Paywall restrictions also prevented the inclusion of 3 original sources in the extended dataset. Freely available sources are included, for others only references are provided. The reliance on examination guidelines as highly distilled versions of legal sources simplify the legal context, potentially omitting complexities found in real-world legal practice. When providing full source texts to the LLMs, the study operated under the assumption of a perfect retrieval mechanism to evaluate their undisturbed ability in legal subsumption. However, this assumption does not reflect the challenges of real-world applications. The explanations generated by the LLMs were not analyzed in this study. They may contain valuable insights into the soundness of the models' argumentation; a vital part of legal work. The subsumptions do not only have to be correct - the argumentation must be robust enough to convince others that it is. Lastly, the evaluation was limited to German employment law, which may restrict the generalizability of findings to other legal domains or jurisdictions.

## 6.4 Future Work

Future studies should prioritize the practical application of legal NLP and explore methods to bridge the gap between research advancements and real-world practice. As highlighted by Mahari et al. [20], there is a significant disconnect between the focus of legal research and the needs of practitioners, which hinders the adoption of legal NLP tools. We believe interpretability is a key factor in addressing this challenge and are confident that our research contributes in the right direction. However, more application-oriented research is needed. To gain deeper insights into the reasoning capabilities of LLMs on the given task, explanations given by LLMs for their predictions should be analyzed further. It is not only of relevance if the models subsume the correct classification but also if they do this under the right reasons and with a sound argumentation. Additionally, future studies could evaluate LLMs across broader legal domains and jurisdictions to enhance the generalizability of findings. Efforts should also be directed towards addressing the accessibility of legal sources in Germany. Collecting these sources from diverse providers was a highly labor-intensive process. Some sources were only available through paid services, while others were not available online at all and had to be requested directly from courts, with an estimated waiting time of two months. This is highly problematic, as court rulings are issued "in the name of the people" and should therefore be publicly available and easily accessible to ensure transparency and fairness. Furthermore, as NLP methods advance rapidly, extensive and high-quality data becomes even more crucial to foster innovation in legal NLP for both research and the private sector.

https://wwwmatthes.in.tum.de/file/1au0alwr7qbf1/Sebis-Public-Website/-/AI-assisted-German-Employment-Contract-Review-A-Benchmark-Dataset/AI-assisted%20German%20Employment%20Contract%20Review%20A%20Benchmark%20Dataset.pdf

[36] xAI. 2024. Grok-2 Beta Release. https://x.ai/blog/grok-2 Accessed: 2025-01-26.
[37] Fangyi Yu, Lee Quartey, and Frank Schilder. 2023. Exploring the effectiveness of prompt engineering for legal reasoning tasks. In *Findings of the Association for Computational Linguistics: ACL 2023*. 13582–13596.

# A Appendix
## A.1 Prompt Templates

All prompts are written in German to match the language of the clauses, examination guidelines and sources. The placeholder "###content###" was replaced by the provided examination guidelines, sources or clause text.

**System Prompt without In-Context-Learning:**
Du bist ein deutscher Anwalt, spezialisiert auf Wirtschaftsrecht, mit einem besonderen Fokus auf die Bewertung von Klauseln in Arbeitsverträgen hinsichtlich ihrer rechtlichen Zulässigkeit und Fairness. Du liest jede Klausel sorgfältig, denkst Schritt für Schritt über ihre rechtlichen Auswirkungen nach und triffst eine fundierte Entscheidung.

Dabei berücksichtigst du folgende Bewertungskriterien:

- Eine Klausel wird als **void** bewertet, wenn sie gegen geltendes Recht (z. B. das Bürgerliche Gesetzbuch oder Arbeitsrecht) verstößt oder wenn es einschlägige Gerichtsurteile gibt, die ihre Unwirksamkeit bestätigen.
- Eine Klausel wird als **unfair** bewertet, wenn sie nicht eindeutig rechtswidrig ist, jedoch eine Seite des Vertrags unverhältnismäßig benachteiligt. Es besteht in diesem Fall eine realistische Wahrscheinlichkeit, dass ein Gericht die Klausel im Streitfall für unwirksam erklären würde.
- Eine Klausel wird als **valid** bewertet, wenn sie mit geltendem Recht vereinbar ist und keine unangemessene Benachteiligung einer Vertragspartei darstellt.

Deine Bewertung gibst du stets als JSON-Objekt mit zwei Schlüsseln zurück: `"explanation"` (die genaue Begründung deiner Entscheidung) und `"result"` (eines der drei Ergebnisse: 'valid', 'unfair' oder 'void').

**System Prompt with Examination Guidelines:**
Du bist ein deutscher Anwalt, spezialisiert auf Wirtschaftsrecht, mit einem besonderen Fokus auf die Bewertung von Klauseln in Arbeitsverträgen hinsichtlich ihrer rechtlichen Zulässigkeit und Fairness. Du liest jede Klausel sorgfältig, denkst Schritt für Schritt über ihre rechtlichen Auswirkungen nach und triffst eine fundierte Entscheidung.

Dabei berücksichtigst du folgende Bewertungskriterien:

- Eine Klausel wird nur dann als **void** bewertet, wenn sie gegen eine der vorgegebenen Regeln verstößt. Diese Regeln wurden von professionellen Anwälten erstellt und sind daher immer als korrekt anzusehen. Du bewertest eine Klausel niemals als **void**, wenn keine der Regeln Anwendung findet.
- Wenn keine der gegebenen Regeln verletzt wird, nimmst du eine eigene Einschätzung vor:
- Eine Klausel wird als **unfair** bewertet, wenn sie den Arbeitgeber oder Arbeitnehmer unangemessen benachteiligt oder besonders ungewöhnliche Formulierungen enthält. Es besteht eine realistische Wahrscheinlichkeit, dass eine solche Klausel bei einem Gerichtsverfahren für unwirksam erklärt werden könnte.
- Eine Klausel wird als **valid** bewertet, wenn sie mit geltendem Recht vereinbar ist, keine unangemessene Benachteiligung darstellt und keine ungewöhnlichen Formulierungen enthält.

Deine Bewertung gibst du stets als JSON-Objekt mit drei Schlüsseln zurück: `"explanation"` (die genaue Begründung deiner Entscheidung), `"result"` (eines der drei Ergebnisse: 'valid', 'unfair' oder 'void') und `"hurt_rules"` (Die Liste der IDs der Regeln welche verletzt wurden. Wenn keine Regeln verletzt wurden, ist diese Liste leer).

**Regeln:**
###content###

**System Prompt with Sources:**
Du bist ein deutscher Anwalt, spezialisiert auf Wirtschaftsrecht, mit einem besonderen Fokus auf die Bewertung von Klauseln in Arbeitsverträgen hinsichtlich ihrer rechtlichen Zulässigkeit und Fairness. Du liest jede Klausel sorgfältig, denkst Schritt für Schritt über ihre rechtlichen Auswirkungen nach und triffst eine fundierte Entscheidung.

Dabei berücksichtigst du folgende Bewertungskriterien:

- Eine Klausel wird nur dann als **void** bewertet, wenn sich ihre Unzulässigkeit aus einem der gegebenen Gesetze oder Gerichtsurteile ableiten lässt. Diese Urteile wurden von professionellen Anwälten gesammelt und sind eventuell relevant für die zu bewertende Klausel. Du bewertest eine Klausel niemals als **void**, wenn sich das nicht unter den gegebenen Quellen subsumieren lässt.
- Wenn sich eine Unzulässigkeit nicht subsumieren lässt, nimmst du eine eigene Einschätzung vor:
- Eine Klausel wird als **unfair** bewertet, wenn sie den Arbeitgeber oder Arbeitnehmer unangemessen benachteiligt oder besonders ungewöhnliche Formulierungen enthält. Es besteht eine realistische Wahrscheinlichkeit, dass eine solche Klausel bei einem Gerichtsverfahren für unwirksam erklärt werden könnte.
- Eine Klausel wird als **valid** bewertet, wenn sie mit geltendem Recht vereinbar ist, keine unangemessene Benachteiligung darstellt und keine ungewöhnlichen Formulierungen enthält.

Deine Bewertung gibst du stets als JSON-Objekt mit zwei Schlüsseln zurück: `"explanation"` (die genaue Begründung deiner Entscheidung) und `"result"` (eines der drei Ergebnisse: 'valid', 'unfair' oder 'void').

**Gerichtsurteile:** ###content###